\documentclass[conference]{IEEEtran}
\IEEEoverridecommandlockouts
\usepackage{cite}
\usepackage{amsmath,amssymb,amsfonts}
\usepackage{algorithmic}
\usepackage{graphicx}
\usepackage{textcomp}
\usepackage{xcolor}
\usepackage{diagbox}
\usepackage{booktabs}
\def\BibTeX{{\rm B\kern-.05em{\sc i\kern-.025em b}\kern-.08em
    T\kern-.1667em\lower.7ex\hbox{E}\kern-.125emX}}
\begin{document}

\title{Long-Short Temporal Co-Teaching \\ for Weakly Supervised Video Anomaly Detection}

\author{\IEEEauthorblockN{Shengyang Sun \quad\quad\quad\quad\quad Xiaojin Gong\textsuperscript{*}\thanks{*Corresponding author.}}
\IEEEauthorblockA{College of Information Science \& Electronic Engineering \\
Zhejiang University, Hangzhou, Zhejiang, China\\
\{sunshy, gongxj\}@zju.edu.cn}
}

\maketitle

\begin{abstract}
	Weakly supervised video anomaly detection (WS-VAD) is a challenging problem that aims to learn VAD models only with video-level annotations. In this work, we propose a Long-Short Temporal Co-teaching (LSTC) method to address the WS-VAD problem. It constructs two tubelet-based spatio-temporal transformer networks to learn from short- and long-term video clips respectively. Each network is trained with respect to a multiple instance learning (MIL)-based ranking loss, together with a cross-entropy loss when clip-level pseudo labels are available. A co-teaching strategy is adopted to train the two networks. That is, clip-level pseudo labels generated from each network are used to supervise the other one at the next training round, and the two networks are learned alternatively and iteratively. Our proposed method is able to better deal with the anomalies with varying durations as well as subtle anomalies. Extensive experiments on three public datasets demonstrate that our method outperforms state-of-the-art WS-VAD methods. Code is available at https://github.com/shengyangsun/LSTC\_VAD.
\end{abstract}
\begin{IEEEkeywords}
	Video anomaly detection, weak supervision, co-teaching strategy
\end{IEEEkeywords}
\vspace{-1pt}
\section{Introduction}
Video anomaly detection (VAD) aims to detect abnormal events in videos. This task has received great research interest because of its significance in security surveillance, accident forecasting, and evidence investigation. A majority of previous studies focus on one class classification paradigm~\cite{cai2021appearance,Sun2023HSC}, which learns a VAD model only using normal videos and identifies those deviating from normal patterns as anomalies. However, owing to the complexity and diversity of real-world scenarios, it is impossible to take into account all normal patterns so these methods are prone to incur false alarms. Recently, weakly-supervised video anomaly detection (WS-VAD) has been also explored. It learns from both normal and abnormal videos annotated with video-level labels. The developed WS-VAD methods~\cite{sultani2018real, feng2021mist, li2022self} demonstrate superior performance than the unsupervised counterparts at the cost of a light annotation burden.

Nevertheless, there are two essential problems that need to be well addressed toward precise WS-VAD. 1) How to leverage video-level labels to identify anomalies in abnormal videos that may be overwhelmed by normal clips? 2)~How to thoroughly capture spatial and temporal contexts that are important for discriminating abnormal events? To address the first problem, various methods have been developed, including multiple instance learning (MIL) based techniques~\cite{sultani2018real}, pseudo labeling together with label noise reducing~\cite{zhong2019graph} or self-training~\cite{feng2021mist,li2022self} techniques. For the second one, pre-trained spatio-temporal feature encoders such as C3D~\cite{tran2015learning} and I3D~\cite{carreira2017quo} are commonly used. Besides, spatio-temporal dual-branch network~\cite{Wu2021}, continuous sampling strategy~\cite{feng2021mist}, and self-attention or transformer techniques~\cite{tian2021weakly,li2022self} have been further applied to capture long-range dependencies. 

In this work, we propose a new WS-VAD method that addresses the above-mentioned problems distinctively. Our method is motivated by the following characteristics observed in abnormal events: 1) The duration of abnormal events varies a lot, lasting for one or multiple consecutive clips. 2)~Abnormal events may take only a small portion of regions in many surveillance videos. Therefore, we choose tubelets (a tube is a clip patch) as tokens and construct two spatio-temporal transformer networks to regress anomaly scores for short- and long-term sequences of clips, respectively. Pseudo labels generated from each network are further used to supervise the other one in the next training round. 

Our method distinguishes itself from the others in the following aspects:
\begin{itemize}
	\item We construct tubelet-based spatio-temporal transformer networks to mine spatial and temporal contexts thoroughly. Although tubelets have become common in recent video transformers~\cite{arnab2021vivit,liu2022video}, most WS-VAD methods~\cite{sultani2018real,tian2021weakly,feng2021mist,li2022self} still exploit clip-level features for anomaly detection. In contrast to them, the tubelet-based way can capture features at a finer granularity so that subtle anomalies can be detected more precisely. 
	\item We employ a co-teaching strategy to train short- and long-term networks alternatively and iteratively. The two networks can explicitly learn from abnormal events with varying durations. Moreover, in contrast to the self-training strategy used in recent WS-VAD methods~\cite{feng2021mist,li2022self}, our co-teaching strategy can better tackle heavy noise in pseudo labels and boost both networks' performance further. 
	\item Experiments on three public datasets demonstrate that the proposed method outperforms state-of-the-art WS-VAD methods. Visualization results show that the short- and long-term networks learned via co-teaching focus on anomalous regions more accurately than the standalone counterparts.
\end{itemize}

\section{Related Work}
\subsection{Weakly Supervised Video Anomaly Detection}
Weakly supervised video anomaly detection learns a VAD model depending on video-level labels. One research line of previous studies formulates this task as a multiple instance learning (MIL) problem. In addition to a MIL ranking loss, various regularization losses have been proposed. For instance, Sultani~\textit{et al.}~\cite{sultani2018real} place sparsity and smoothness constraints, Zhang~\textit{et al.}~\cite{zhang2019temporal} define an inner bag loss, and Wan~\textit{et al.}~\cite{wan2020weakly} design a dynamic MIL loss together with a center loss to regularize the learned feature space. Meanwhile, another research line formulates WS-VAD as a supervised learning task that takes video-level labels directly as clip-level pseudo labels. To mitigate the label noise contained in abnormal videos, Zhong~\textit{et al.}~\cite{zhong2019graph} propose a graph convolutional network and Zaheer~\textit{et al.}~\cite{zaheer2020claws} define a clustering based loss. 

Recently, the two research lines are combined in several state-of-the-art works~\cite{feng2021mist,li2022self}. They first construct a VAD model based on the MIL framework to produce clip-level pseudo labels and then use the generated labels to supervise the subsequent training of the VAD model. The noisy pseudo labels and the VAD model are refined alternatively and iteratively through a self-training strategy. However, simply utilizing pseudo labels is not a highly effective strategy for datasets that have various durations of anomalies, such as UCF-Crime. Therefore, we propose a co-teaching strategy to alternatively learn two VAD models, by which noisy pseudo labels are cleaned more effectively and the discrimination abilities of both models are boosted considerably.

\subsection{Self-training and Co-training Strategies}
The self-training strategy has been recently applied to unsupervised~\cite{pang2020self, Yu2022} and weakly supervised~\cite{feng2021mist,li2022self} VAD tasks. It alternates between a pseudo-labeling step and a VAD model training step to iteratively promote the model's performance. In self-training, only a single model is maintained, it may suffer from performance degradation on large-scale datasets that have varied durations anomalies. Contrastively, co-training, also known as co-teaching~\cite{Han2018Coteaching}, is a new learning paradigm that trains two models simultaneously and lets them teach each other. Due to different learning and sample-selection bias, the co-trained models are more robust to severe label noise. The co-teaching strategy has been recently applied to unsupervised person re-identification~\cite{Yang2020,Ge2020MMT}, object detection~\cite{Wang2020coteaching}, and other vision tasks. In this work, we introduce the co-teaching strategy to WS-VAD, by which a short-term network and a long-term network can learn from each other.

\subsection{Spatial-Temporal Contexts in WS-VAD}
The exploitation of spatial and temporal contexts plays an important role in VAD. Many WS-VAD methods exploit both information through pre-trained spatio-temporal feature extractors such as C3D~\cite{tran2015learning} and I3D~\cite{carreira2017quo}. Besides, long-range temporal dependencies have been further captured, \textit{e.g.} Tian~\textit{et al.}~\cite{tian2021weakly} employ the temporal dilated convolution and Li~\textit{et al.}~\cite{li2022self} employ a transformer-based multiple-sequence learning technique. The synergy of long-range spatial and temporal dependencies has been explored, \textit{e.g.} Purwanto~\textit{et al.}~\cite{purwanto2021dance} employ the self-attention and conditional random field, and Wu~\textit{et al.}~\cite{Wu2021} propose a spatio-temporal dual-branch network architecture. However, all above previous works have not considered both fine-grained spatial features and long-range temporal dependencies. In this work, we construct two tubelet-based spatio-temporal transformer networks to learn from short- and long-term clips respectively and adopt the co-teaching strategy to enhance both networks. By this means, fine-grained features as well as spatial and long-range temporal dependencies are all captured.

\begin{figure*}[t]
	\centering
	\includegraphics[width=0.95\textwidth]{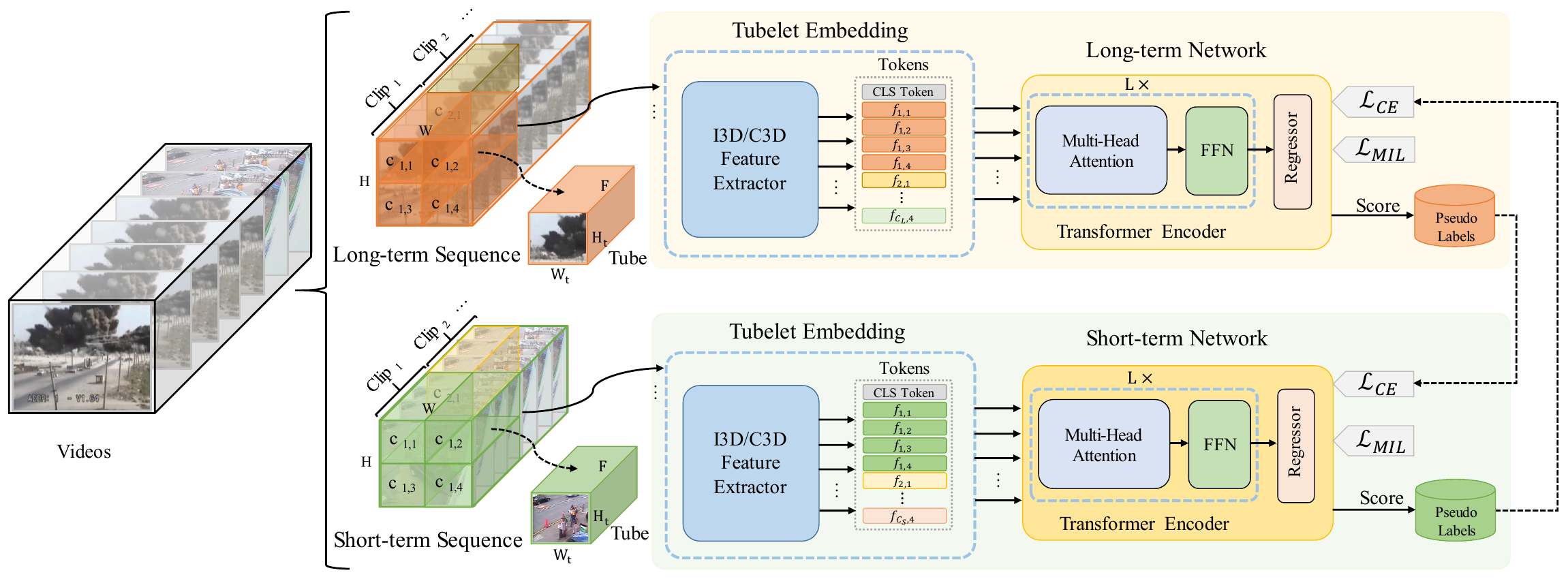} 
	\caption{An overview of the proposed method. It adopts a co-teaching strategy to train a short-term network and a long-term network alternatively. Each network takes short- or long-term clips as input. The input clips are first split into tubelets (taking 4 tubes as an example) and passed through a pre-trained feature extractor for tokenization. The embedded tokens are further input into a spatio-temporal transformer network composed of multiple transformer layers and a regression layer to predict anomaly scores. Pseudo labels generated from the scores predicted by one network are used to supervise the other network. 
	}
\label{fig:SAT}
\end{figure*}

\section{The Proposed Method}
Fig.~\ref{fig:SAT} presents an overview of the proposed method. In this section, we first introduce the transformer-based short- or long-term network architecture, then present the training losses and the co-teaching strategy.

\subsection{Transformer-based VAD Network}
To mine fine-grained features as well as capture long-range dependencies, we construct a tubelet-based spatio-temporal transformer as our VAD network architecture. It first embeds tubelet tokens and then passes the tokens through multiple transformer layers and one regression layer to predict an anomaly score for all input clips.

\noindent\textbf{Tubelet Embedding.} Let us consider a clip of dimension $F\times H\times W$, where $F$ is the number of consecutive frames, $H$ and $W$ are the height and width of each frame. We split the clip into non-overlapping tubes as shown in Fig.~\ref{fig:SAT}. Each tubelet is of dimension $F \times H_t \times W_t$ and therefore $N_t = \lfloor \frac{H}{H_t}\rfloor \cdot \lfloor \frac{W}{W_t}\rfloor$ tubelets are obtained. We feed each tubelet into a pre-trained feature encoder such as C3D or I3D to extract a $d$-dimensional feature $\mathbf{f} \in\mathbb{R}^d$, which is taken as a token for subsequent transformer layers. In contrast to clip-level tokens used in previous WS-VAD methods~\cite{tian2021weakly,feng2021mist,li2022self}, tubelets capture spatial contexts at a finer granularity, benefiting the detection of subtle anomalies. 
Meanwhile, in contrast to the tokenization way that linearly projects each tubelet into 1D token as in recent video transformers~\cite{arnab2021vivit}, the tubelet feature extracted by C3D or I3D can infuse inductive bias which facilitates the training of transformer networks.

\noindent\textbf{Transformer Architecture.}
Our transformer-based VAD network takes $C$ consecutive clips as input. The clips are tokenized by the above-mentioned way to get a sequence of tokens. Then, a token $\mathbf{z}_{cls}\in\mathbb{R}^d$ is prepended to the sequence to form the input of the transformer. That is, the input sequence is:
\begin{equation}
	\mathbf{z} = [\mathbf{z}_{cls}, \mathbf{f}_{1,1}, ..., \mathbf{f}_{i,j}, ..., \mathbf{f}_{C, N_t}],
\end{equation}
in which $\mathbf{f}_{i,j}$ denotes the $j$-th token in the $i$-th clip. 

The input sequence is passed through a transformer network composed of $L$ transformer layers and one anomaly regression layer. Each transformer layer consists of Multi-head Self-Attention (MSA) and FeedForward Network (FFN) modules, which are respectively implemented by scaled dot-product attention and a 2-layer MLP with ReLU in between. The computation of layer $l$ is as follows:
\begin{equation} \label{eq:MSA_FFN}
	\begin{aligned}
		\hat{\mathbf{z}}^{l} &= \text{MSA}(\text{LN}(\mathbf{z}^{l})) + \mathbf{z}^{l}, \\
		\mathbf{z}^{l+1} &= \text{FFN}(\text{LN}(\hat{\mathbf{z}}^{l})) + \hat{\mathbf{z}}^{l},
	\end{aligned}
\end{equation}
in which LN denotes layer normalization. Finally, the anomaly regressor, which is implemented by a 3-layer MLP with ReLU and Sigmoid activations, takes the output $\mathbf{z}^L_{cls}$ of the last transformer layer to predict an anomaly score $s \in (0, 1)$ for all the clips.
Considering that it is beneficial to encode spatial and temporal position information, we include 3D relative position bias to each head in self-attention computation in MSA, as in~\cite{liu2022video}. That is,
\begin{equation}
	\text{Attention}(\mathbf{Q}, \mathbf{K}, \mathbf{V}) = \text{SoftMax}(\frac{\mathbf{Q}\mathbf{K}^{T}}{\sqrt{d}} + \mathbf{B}) \mathbf{V},
\end{equation}
in which $\mathbf{Q}, \mathbf{K}, \mathbf{V} \in \mathbb{R}^{C\cdot N_t \times d}$ are query, key and value matrices, $C \cdot N_t$ is the number of tokens, and $\mathbf{B} \in \mathbb{R}^{C^2 \times N_t \times N_t}$ is the 3D relative position bias for position encoding. 

\subsection{Training Losses of the VAD Network}
We adopt a MIL ranking loss to train the VAD networks. In addition, since our work employs co-teaching to train two VAD models, in which one produces pseudo labels to supervise the other in the next training round, we therefore also adopt a cross-entropy loss when pseudo labels are available.

\noindent\textbf{MIL Ranking Loss.} The MIL ranking loss is extensively adopted in previous methods~\cite{sultani2018real,feng2021mist,li2022self} that formulate WS-VAD as a multiple instance learning problem. Specifically, it considers a video set $\mathcal{V}=\{V_i\}^{N}_{i=1}$, together with a video-level label set $\mathcal{Y}=\{Y_i\in \{0,1\}\}^{N}_{i=1}$ indicating whether a video is abnormal ($Y_i=1$)  or not ($Y_i=0$). Each video $V_i$ contains $N_i$ clips and each clip has $F$ frames. The MIL-based formulation treats each video as a bag and each clip as an instance. Therefore, an abnormal video is a positive bag $\mathcal{B}^a=\{c^a_j\}^{N_i}_{j=1}$ containing at least one abnormal clip, while a normal video is a negative bag $\mathcal{B}^n=\{c^n_j\}^{N_i}_{j=1}$ having no abnormal instances.  

Following~\cite{feng2021mist, li2022self}, we adopt a continuous sampling strategy and a hinge-based MIL ranking loss to train our VAD network. $K$ subsets of clips are uniformly sampled from each video, and each subset contains $C$ consecutive clips, in which $C$ is the number of clips input to the VAD network. For each subset $k$, our network predicts one anomaly score $s_k$. MIL-based methods assume that the highest anomaly score of subsets from a positive bag is higher than that from a negative bag, that is $\mathop{max}\limits_{1\leq k\leq K}s^{a}_{k} > \mathop{max}\limits_{1\leq k\leq K}s^{n}_{k}$. Then, given a positive bag and a negative bag, we define a hinge-based ranking loss with sparse regularization as follows:
\begin{equation} \label{eq:loss_deep_MIL}
	\small
	\mathcal{L}_{MIL}(\mathcal{B}^a, \mathcal{B}^n)=\left(\tau-\mathop{max}\limits_{1\leq k\leq K}s^{a}_{k}+\mathop{max}\limits_{1\leq k\leq K}s^{n}_{k}\right)_{+}+\frac{\alpha}{K}\sum\limits^{K}_{k=1} s^{a}_k,
\end{equation}
where $(\cdot)_{+}$ denotes $max(0, \cdot)$. The first term is to keep a margin $\tau$ between positive and negative subsets, and the second term is to regularize that only a few subsets contain anomalies. $\alpha$ is a hyper-parameter to balance two terms.

\noindent\textbf{Cross Entropy Loss.}
When clip-level pseudo labels are available (the way of generating pseudo labels will be introduced in Fig.~\ref{sec:co-teaching}), we treat WS-VAD as a supervised learning problem and adopt the commonly used cross-entropy loss for learning. Given the predicted anomaly score $s_i$ of clip $c_i$ and its pseudo label $\tilde{y}_i$, the loss is defined as follows:
\begin{equation} \label{eq:loss_CE}
	\mathcal{L}_{CE}(s_i, \tilde{y}_i) = -\tilde{y}_i log(s_i)-	(1-\tilde{y}_i)log(1-s_i).
\end{equation}

\begin{table*}[tbp]
	\centering
	\caption{The AUC (\%) performance of the short- and long-term networks obtained with respect to the different number of input clips. In each entry, the left AUC is obtained by STN, and the right one is obtained by LTN.}
	\resizebox{.95\textwidth}{!}{
		\begin{tabular}{c|ccccccc}
			\toprule
			\diagbox[dir=NW]{STN}{LTN} & 1 & 2 & 3 & 4 & 5 & 6 & 7 \\
			\midrule
			1 & 96.83/96.98 & 96.93/97.45 & \textbf{97.00}/\textbf{97.79} & 96.87/97.60 & 96.84/97.01 & 96.84/96.94  & 96.84/96.71 \\ 
			2 & - & 95.55/95.68 & 95.68/95.46 & 95.85/96.23 & 95.31/95.37 & 95.31/95.15 & 95.31/95.09 \\ 
			3 & - & - & 96.49/96.54 & 96.54/96.41 & 96.54/96.08 & 96.54/95.66 & 96.54/95.74 \\
			4 & - & - & - & 96.04/96.07 & 96.52/96.27 & 96.26/95.82 & 96.07/95.25\\
			5 & - & - & - & - & 96.21/96.42 & 96.15/95.98 & 96.08/95.32\\
			6 & - & - & - & - & - & 96.35/96.41 & 96.05/95.21 \\
			7 & - & - & - & - & - & - & 95.83/95.92 \\
			\bottomrule
		\end{tabular}
	}
	\label{tab:number_of_clips}
\end{table*}

\subsection{Long-Short Temporal Co-teaching}
\label{sec:co-teaching}
To deal with the varied duration of abnormal events, we construct a short-term network (STN) and a long-term network (LTN) for learning. The former takes a short sequence of clips as input, while the latter inputs a long clip sequence. The two networks are alternatively trained via a co-teaching strategy. In the first round, the STN model is initially trained using the MIL loss $\mathcal{L}_{MIL}$ since no clip-level pseudo labels are available. Later on, once a network is trained, we generate pseudo labels based on its predicted anomaly scores as follows:
\begin{equation} \label{eq:clip_level_score}
	\tilde{y}_i =\left\{
	\begin{aligned}
		s_i &, \quad \text{if} \ \ s_i > \mu\\
		0 &, \quad \text{if} \ \ s_i \leq \mu \ \ \text{or} \ \ Y_i=0,\\
	\end{aligned}
	\right.
\end{equation}
where $\mu$ is a threshold and $Y_i$ indicates the video-level label of clip $i$. Then, the other network takes the pseudo labels for supervision and uses the following loss for training:
\begin{equation} \label{eq:loss_SAP}
	\mathcal{L} = \mathcal{L}_{MIL} + \beta\mathcal{L}_{CE},
\end{equation}
where $\beta$ is a hyper-parameter for balancing two losses.

\subsection{Inference}
During the test, we simply choose the model that performed better on training sets as the final model for inference.

\section{Experiments}
\subsection{Datasets and Evaluation Metrics}
\noindent\textbf{Datasets.} We evaluate the proposed method on three public datasets: ShanghaiTech~\cite{Luo2017ShanghaiTech,zhong2019graph}, UCF-Crime~\cite{sultani2018real}, and UBnormal~\cite{Acsintoae_2022_CVPR}. ShanghaiTech~\cite{Luo2017ShanghaiTech} consists of 437 videos with 130 abnormal events in 13 scenes. In the original dataset, all training videos are normal. \cite{zhong2019graph} reorganized the dataset by selecting some abnormal testing videos into the training set for WS-VAD. We follow this new split way in our experiments. UCF-Crime~\cite{sultani2018real} contains 1,900 untrimmed videos with 13 different types of anomalies, among which 1,610 videos are for training and 290 videos are for the test. UBnormal~\cite{Acsintoae_2022_CVPR} is a new large-scale dataset synthesized by the Cinema4D software. It consists of 543 videos with 22 types of anomalies on 29 virtual scenes, including 268 videos for training, 64 videos for validation, and 211 videos for test. Since it has a higher number of anomaly types occurring across a larger set of scenes, this dataset is more challenging for anomaly detection.

\noindent\textbf{Evaluation Metrics.} Following the common practice~\cite{feng2021mist, li2022self}, we take the area under the curve (AUC) of the frame-level receiver operating characteristics (ROC) for evaluation. A higher AUC indicates better performance. 

\subsection{Implementation Details}
In tubelet embedding, we use pre-trained C3D/I3D to extract features for each tubelet, and both extractors are pre-trained on Kinetics-400~\cite{Kay2017} and directly used on ShanghaiTech and UCF-Crime. Following~\cite{Acsintoae_2022_CVPR}, feature extractors are fine-tuned for 20 epochs on UBnormal. Our transformer networks have 3 layers and the number of heads in MSA is 8. In MIL ranking loss, $K=16$ subsets of clips are sampled for each video on ShanghaiTech and UBnormal, and $K=32$ for UCF-Crime because the videos in this dataset are much longer. The number of tubelets per clip is set to $4\times 4$, $3\times 3$, and $5\times 5$ for ShanghaiTech, UCF-Crime, and UBnormal respectively, according to their video resolutions. The hyper-parameters involved in losses are set as follows: $\alpha = 0.01$ and $\beta = 0.8$. The margin $\tau = 1$, and the score threshold $\mu = 0.85$. The total number of rounds in co-teaching is $R = 4$. In addition, our models are trained using the AdaGrad optimizer with a batch size of 40, a learning rate of 0.0001 for the transformer layers, and 0.01 for the anomaly regressor.

\begin{figure*}[t]
	\centering
	\includegraphics[width=1\textwidth]{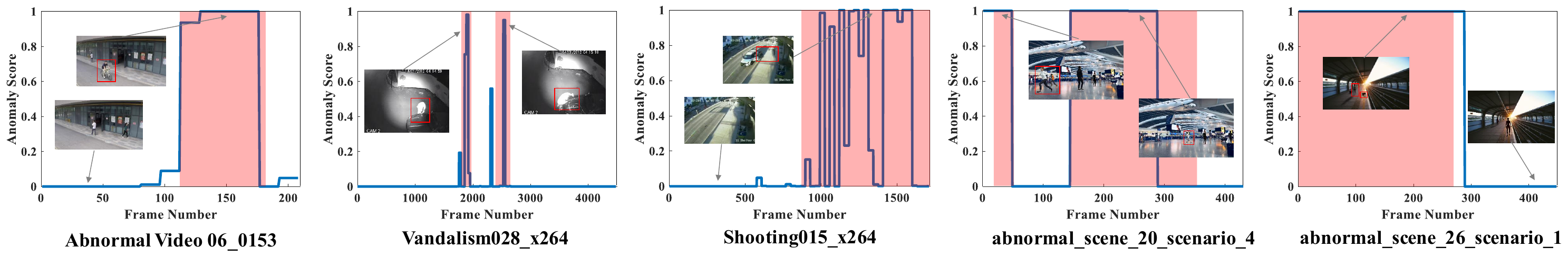}
	\caption{Visualization of anomaly scores predicted on ShanghaiTech (06\_0153), UCF-Crime (Vandalism028\_x264, Shooting015\_x264) and UBnormal (20\_scenario\_4, 26\_scenario\_1). The blue lines are predicted anomaly scores and the red regions indicate ground-truth events.}
	\label{fig:visual_figure}
\end{figure*}

\subsection{Ablation Studies}
We first conduct a series of experiments to validate the effectiveness of the proposed method and analyze the impact of critical hyper-parameters. All experiments in ablation studies use the I3D-RGB feature encoder and one-crop setting.

\noindent\textbf{The Impact of Input Clip Number.} In the proposed method, we input a different number of clips into the short- and long-term networks to learn abnormal events with varying durations. 
We hereby investigate the impact of the input clip number by varying the number from 1 to 7 for STN and LTN. When the clip number is 1, we adopt the continuous sampling strategy to select $T=7$ clips for each subset and get one anomaly score for each subset via average pooling as in~\cite{feng2021mist}. Fig.~\ref{tab:number_of_clips} presents the corresponding results achieved on ShanghaiTech. Roughly speaking, the performance first increases and then gradually drops when the clip number goes up for both networks. The best performance is obtained when the STN model takes one clip as input and the LTN model inputs three clips. Therefore, we fix these numbers throughout all the following experiments and adopt LTN for inference. 

\begin{table}[t]
	\centering
	\caption{Comparison of frame-level AUC performance with other SOTA methods. $\dagger$ indicates this result is reported by \cite{feng2021mist}, \cite{tian2021weakly} or \cite{Acsintoae_2022_CVPR}, and * indicates we re-train this method using official implementations. The best result is bolded.}
	\resizebox{.49\textwidth}{!}{
		\begin{tabular}{c|c|c|c|c|c}
			\toprule
			Method & Feature & Crop & SHT & UCF & UBnormal\\
			\midrule
			MIL \cite{sultani2018real} & C3D-RGB & one & 86.30$\dagger$ & 75.41 & 50.30$\dagger$\\
			MIL \cite{sultani2018real} & I3D-RGB & ten & 85.33$\dagger$ & 77.92$\dagger$ & 62.05* \\
			GCN \cite{zhong2019graph} & C3D-RGB & ten & 76.44 & 81.08 & - \\
			MIST \cite{feng2021mist} & C3D-RGB & one & 93.13 & 81.40 & 59.57*\\
			MIST \cite{feng2021mist} & I3D-RGB & one & 94.83 & 82.30 & 68.21*\\
			RTFM \cite{tian2021weakly} & C3D-RGB & ten & 91.51 & 83.28 & 64.74*\\
			RTFM \cite{tian2021weakly} & I3D-RGB & ten & 97.21 & 84.30 & 69.41*\\
			MSL \cite{li2022self} & C3D-RGB & one & 94.23 & 82.85 & -\\
			MSL \cite{li2022self} & C3D-RGB & ten & 94.81 & - & -\\
			MSL \cite{li2022self} & I3D-RGB & one & 95.45 & 85.30 & -\\
			MSL \cite{li2022self} & I3D-RGB & ten & 96.08 & - & -\\
			MSL \cite{li2022self} & VS-RGB & one & 96.93 & 85.62 & -\\
			MSL \cite{li2022self} & VS-RGB & ten & 97.32 & - & -\\
			\midrule
			LSTC (This work) & C3D-RGB & one & \textbf{95.47} & \textbf{83.10} & \textbf{65.26}\\ 
			LSTC (This work) & C3D-RGB & ten & \textbf{96.56} & \textbf{83.47} & \textbf{66.28}\\ 
			LSTC (This work) & I3D-RGB & one & \textbf{97.79} & \textbf{85.70} & \textbf{77.51}\\ 
			LSTC (This work) & I3D-RGB & ten & \textbf{97.92} & \textbf{85.88} & \textbf{77.92}\\
			\bottomrule
		\end{tabular}
	}
	\label{tab:compare_auc}
\end{table}

\begin{figure}[b]
	\vspace{-8pt}
	\centering
	\includegraphics[width=1\columnwidth]{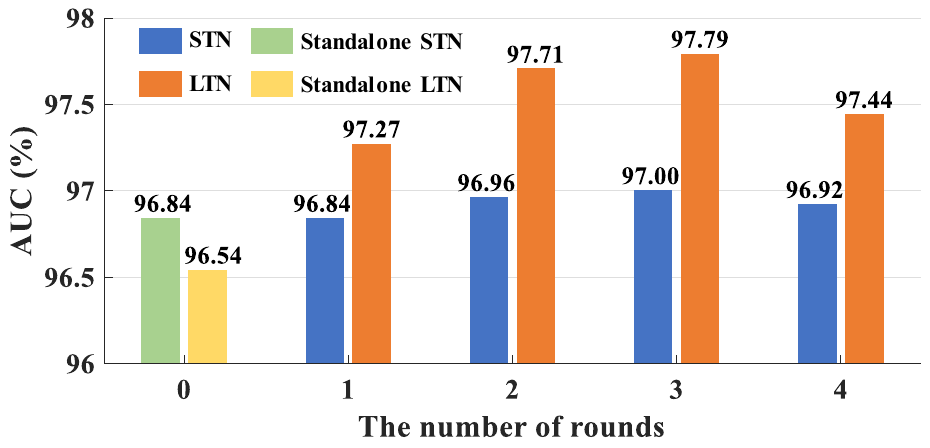} 
	\caption{The performance of standalone and co-trained models at different rounds.}
	\label{fig:quantitative_comparison}
\end{figure}

\noindent\textbf{The Impact of the Co-teaching Strategy.} The LTN model and the STN model have been trained alternatively in each round within the co-teaching strategy. We set the total number of rounds to be 4, which is enough for our models to converge on all datasets. To validate the effectiveness of the co-teaching strategy, we compare the models trained via co-teaching with the standalone models that are individually trained with $\mathcal{L}_{MIL}$. Fig.~\ref{fig:quantitative_comparison} presents the performance of standalone models, together with the performance of co-trained models in different rounds on ShanghaiTech. The results show that both co-trained models gradually improved in the first 3 rounds and then degenerate in the last round due to the overfitting to noise. The best models are obtained at round 3, in which LTN and STN outperform their standalone counterparts by $1.25\%$ and $0.16\%$, respectively, validating the effectiveness of the co-teaching strategy.

\subsection{Comparison to State-of-the-Art}
We compare our method with state-of-the-art WS-VAD methods in Table~\ref{tab:compare_auc}. In experiments, both C3D-RGB and I3D-RGB feature extractors, together with one-crop and ten-crop data augmentations, are investigated.

\noindent\textbf{Results on ShanghaiTech.} Our method outperforms all previous methods when conducted in the same settings. 
Specifically, our best performed model achieves $97.92\%$ AUC, which is $0.6\%$ higher than the second placed method (i.e. MSL). The superior performance is achieved by both the fine-grained transformer architecture and the co-teaching strategy. Note that the MSL method with the VideoSwin-RGB (VS-RGB) feature encoder also captures fine-grained features, but our method still outperforms it by a considerable margin. 

\noindent\textbf{Results on UCF-Crime.} With the I3D-RGB feature encoder and one-crop augmentation, the proposed method achieves $85.70\%$ AUC, higher than all other methods even if they use a ten-crop or the more advanced VideoSwin-RGB (VS-RGB) encoder. In addition, our best performed model achieves $85.88\%$ AUC, which is $0.26\%$ higher than the second placed method (i.e. MSL).

\noindent\textbf{Results on UBnormal.} Since this dataset is released very recently, we can not get the published performance for most SOTA methods except those reported in~\cite{Acsintoae_2022_CVPR}. Therefore we re-train the recent methods if their codes are available and report their performance. From the results, we see that the proposed method outperforms all other methods by a great margin. 

\begin{figure}[t]
	\centering
	\includegraphics[width=0.48\textwidth]{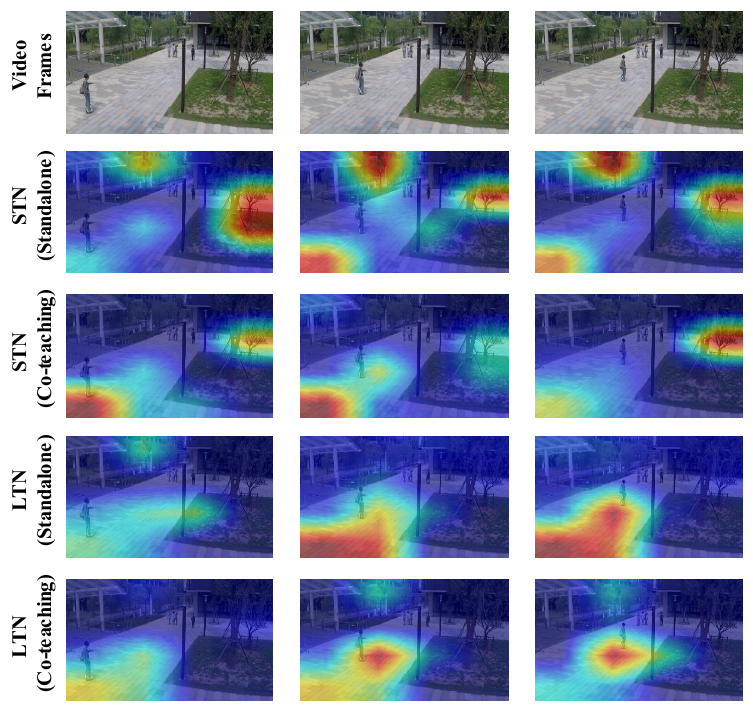} 
	\caption{Visualization of anomaly attention maps via the Attention Rollout scheme~\cite{bertasius2021space} on ShanghaiTech (08\_0157). }
	\label{fig:visual_attention}
\end{figure}

\subsection {Qualitative Visualization}

\noindent\textbf{Predicted Anomaly Scores.} Fig.~\ref{fig:visual_figure} illustrates the predicted anomaly scores for testing videos on three datasets. Various anomaly events across diverse scenes are presented. As shown in the figure, the proposed method can detect short- (e.g. Vandalism028\_x264) and long-term (e.g. Shooting015\_x264 and abnormal\_scene\_26\_scenario\_1) abnormal events precisely. It can also detect multiple abnormal events in one video (e.g. Vandalism028\_x264 and abnormal\_scene\_20\_scenario\_4). Moreover, our method predicts anomaly scores very close to 1 for most abnormal events and very close to 0 for normal videos. 

\noindent\textbf{Anomaly Attention Maps.} 
We visualize anomaly attention maps via the Attention Rollout scheme~\cite{bertasius2021space, abnar2020quantifying} in Fig.~\ref{fig:visual_attention}. It can be observed that the STN and LTN models learned by co-teaching attend to abnormal regions more accurately than the standalone counterparts. Moreover, the STN model focuses more on abnormal objects while the LTN model is able to attend to the trajectory of abnormal motions. Such complementary abilities help them to boost each other during the co-teaching procedure.

\section{Conclusion}
In this work, we have presented a long-short temporal co-teaching (LSTC) method to address the WS-VAD problem. It constructs two tubelet-based spatio-temporal transformer networks to learn from short- and long-term video clips respectively. The two models are trained alternatively and iteratively via a co-teaching strategy. Benefiting from the tubelet-based transformer architecture and the long-short temporal co-teaching strategy, the proposed method is able to better deal with subtle anomalies and anomalies with varying durations. Experiments show that our method performs better than state-of-the-art methods.

\bibliographystyle{IEEEtran}
\bibliography{icme2023_LSTC}

\end{document}